\documentclass[runningheads]{llncs}

\usepackage[T1]{fontenc}
\usepackage{graphicx}
\usepackage{xcolor}
\usepackage{amsmath}
\usepackage{booktabs}
\usepackage{tabularx}
\usepackage{float}
\usepackage{tabularray}
\UseTblrLibrary{booktabs}
\usepackage{url}
\usepackage{fvextra}

\DefineVerbatimEnvironment{PromptBlock}{Verbatim}{%
  fontsize=\scriptsize,
  breaklines=true,
  breakanywhere=true,
  frame=single,
  framesep=3pt,
  framerule=0.3pt
}

\begin{document}
\nocite{*}

\title{Agentic Data Sampling Framework for Multilingual ASR via Linguistic-Aware Error-Driven Optimization}
\titlerunning{Agentic Data Sampling for Multilingual ASR}
\author{Zhiheng Qian\inst{1}\fnmsep\thanks{Equal contribution. $^\dagger$Corresponding author.} \and
Junhao Du\inst{2,\star} \and
Jing Peng\inst{1} \and
Hanqi Li\inst{1} \and
Kunyang Peng\inst{2} \and
Feng Xue\inst{2} \and
Lu Chen\inst{1} \and
Shuai Wang\inst{3} \and
Kai Yu\inst{1,\dagger}}
\authorrunning{Z. Qian et al.}
\institute{Shanghai Jiao Tong University
\and
AISpeech Co., Ltd.
\and
Nanjing University\\[0.5em]
\email{\{n1vnhil,jing.peng,daqige,chenlusz,kai.yu\}@sjtu.edu.cn}\\
\email{dujunhao0114@gmail.com}\\
\email{\{kunyang.peng,feng.xue\}@aispeech.com}\\
\email{shuaiwang@nju.edu.cn}}

\maketitle

\begin{abstract}

Multilingual automatic speech recognition (ASR) has achieved rapid progress with large-scale pretrained models and multilingual speech corpora. 
Previous studies have proved the pivotal role of data sampling in multilingual ASR performance. However, existing sampling strategies largely rely on time-consuming manual, heuristic-based designs.
To address this, we propose a linguistic-aware agentic data sampling framework that introduces a data sampling LLM agent into the multilingual ASR training loop. The agent is driven by a linguistically grounded rubric that analyzes ASR errors from phonological, morphological, and syntactic perspectives, and uses the resulting feedback together with training statistics, budget constraints, and algorithm profiles to select domain-level, sample-level, and token-level sampling strategies. This design turns evaluation results into adaptive data-selection decisions rather than treating WER/CER as final endpoints only. We conduct experiments on twelve Common Voice languages using Whisper-medium as the ASR backbone. The results show that our method achieves performance better than full-data training while using less training data and lower computational resources, and substantially accelerates training convergence compared to conventional sampling strategies.

\end{abstract}

\section{Introduction}

Multilingual automatic speech recognition (ASR) aims to recognize
multiple languages within a unified model, reducing the cost of
developing and maintaining language-specific systems. Large-scale joint
training has demonstrated the value of cross-lingual parameter sharing,
particularly for low-resource languages~\cite{massively_multilingual_asr}.
Cross-lingual self-supervised pretraining, weakly supervised
sequence-to-sequence learning, and massively multilingual scaling have
further extended language coverage and transferability
~\cite{xlsr,whisper,mms}. Despite this progress, recognition performance
remains uneven across languages because multilingual corpora differ
substantially in data availability, acoustic conditions, pronunciation
systems, morphology, and writing conventions.

A central challenge is therefore how to use heterogeneous multilingual
training data effectively under a limited budget. Existing approaches
typically concatenate corpora, balance languages, follow predefined
curricula, or select auxiliary data according to corpus relatedness,
historical feedback, or acoustic and phonetic similarity
~\cite{corpus_relatedness,adversarial_meta_sampling,dynamic_sampling_meta_asr,acoustic_pseudotokens}. These methods
demonstrate that multilingual data are not equally useful. However, most
optimize a predefined criterion at a single granularity, such as
language weighting or utterance selection. They do not jointly adapt
language allocation, utterance selection, and token supervision to the
model's evolving error profile.

Current evaluation practice provides limited support for such adaptation.
Multilingual benchmarks commonly report average and per-language WER/CER~\cite{mlsuperb}. Although essential for system comparison, these
metrics are affected by tokenization, writing systems, morphology, and
normalization~\cite{cer_multilingual,normalization_multilingual}, and do
not identify the linguistic source of an error. Typological studies show
that phonological, morphological, and syntactic properties help explain
cross-language performance differences~\cite{typology_multilingual_asr}.
Nevertheless, such analyses are generally performed after training and
are rarely converted into decisions about what data the model should
learn from next. This leaves a gap between multilingual ASR error
diagnosis and data optimization.

In this work, we propose a linguistic-aware agentic data sampling
framework that closes this loop. After each training round, recognition
outputs are analyzed through a linguistic rubric covering phonology,
morphology, and syntax. An LLM agent combines these diagnostics with
per-language performance, training history, budget constraints, and
algorithm profiles. It then produces an executable sampling plan
consisting of domain-, sample-, and token-level algorithm choices and
their parameters for the next round. Data sampling is thus formulated as
an adaptive, feedback-driven process rather than a fixed preprocessing
strategy.

Our contributions are threefold. First, we formulate multilingual ASR
data sampling as a closed-loop optimization problem in which evaluation
results directly guide subsequent training-data selection. Second, we
provide a linguistic error rubric and a hierarchical agentic sampler
that coordinate language allocation, utterance selection, and
token-level supervision according to the current model state. Third,
extensive experiments on twelve Common Voice languages show that the
proposed method outperforms both no-sampling and random-sampling
baselines in average multilingual recognition performance under the
same cumulative training budget, while achieving the best results on
most evaluated languages. Further analyses demonstrate improved
training convergence, consistent reductions in phonological,
morphological, and function-word errors, and adaptive changes in the
agent's decisions as the model error profile evolves.

\section{Related Works}
\subsection{Evaluation and Error Diagnosis in Multilingual ASR}
\label{sec:multilingual_asr_evaluation}

Multilingual ASR evaluation has developed from testing individual
languages on separate corpora toward broader and more standardized
cross-lingual benchmarks. Common Voice provides diverse
community-contributed speech across many resource conditions, while
FLEURS enables more controlled comparison through parallel content
across languages~\cite{commonvoice,fleurs}. XTREME-S and the
ML-SUPERB series further evaluate multilingual speech representations
across languages, datasets, adaptation settings, and related speech
tasks~\cite{mlsuperb,xtremes,mlsuperb2}. Recent extensions additionally
consider language varieties, accents, dialects, code-switching, and
long-form recognition~\cite{mlsuperb_challenge,csfleurs,open_asr_leaderboard}. These benchmarks have made per-language
performance and cross-lingual robustness increasingly visible, rather
than relying only on a single macro-average score.

Despite this broader coverage, multilingual ASR is still evaluated
primarily with WER and CER. These metrics provide reproducible measures
of transcription accuracy, but their interpretation varies with writing
systems, tokenization conventions, word boundaries, and morphological
structure. CER can offer greater consistency than WER for some
multilingual comparisons, but neither metric identifies the linguistic
source of an error~\cite{cer_multilingual}. Reported results are also
sensitive to evaluation design: language-insensitive normalization may
remove meaningful orthographic information, while annotation and
language-label inconsistencies in multilingual corpora can distort the
measured performance~\cite{normalization_multilingual,data_quality_multilingual}.

This limitation is important because similar WER/CER values may reflect
different model weaknesses. Errors may arise from phone confusions and
phonotactic patterns, missing inflectional markers, or incorrectly
recognized function words. Typologically informed analyses have shown
that phonological, morphological, and syntactic properties help explain
cross-language performance variation in multilingual ASR
systems~\cite{typology_multilingual_asr}. Such analyses therefore offer
a more informative view of model behavior than aggregate edit distance
alone.

However, existing evaluation and diagnostic studies are mainly used for
benchmarking or post-hoc interpretation. They reveal which languages or
linguistic properties are difficult, but usually do not determine how
the training data should be changed in response. This leaves a gap
between error diagnosis and model optimization. Our work addresses this
gap by converting phonological, morphological, and syntactic error
statistics into structured feedback for subsequent domain-, sample-,
and token-level data selection. In this way, multilingual evaluation is
used not only to measure the final model, but also to guide the next
training round.
\subsection{Data Sampling Algorithms}
\label{sec:data_sampling_algorithms}

Data sampling and selection improve training efficiency by assigning different importance to data sources, examples, or supervision units.
At the domain level, DoReMi learns mixtures of heterogeneous data
sources using proxy-model excess losses and distributionally robust
optimization~\cite{doremi}. DoGE estimates the contribution of each
domain to target generalization through bilevel optimization and
gradient-based signals~\cite{doge}, while exponentiated-gradient
reweighting provides a general mechanism for converting domain scores
into adaptive sampling distributions~\cite{exp_reweight}. These methods
show that a fixed data mixture may become suboptimal as training
progresses.

Finer-grained methods operate on individual examples or tokens. LESS
uses low-dimensional gradient representations to select examples that
align with a target or validation objective~\cite{less}. RHO-1
prioritizes under-learned tokens according to their excess loss relative
to a reference model~\cite{rho1}, whereas TOSS uses token-level risk
signals to suppress undesirable supervision~\cite{toss}. Although these
methods were mainly developed for language-model training, they
establish complementary mechanisms for domain reweighting, example
selection, and token-level control.

Multilingual ASR introduces additional selection criteria because the
value of speech data depends on both linguistic and acoustic
relationships. Corpus-relatedness sampling selects auxiliary speech
according to source--target similarity~\cite{corpus_relatedness}, while
adversarial meta-sampling adapts language selection using historical
model feedback~\cite{adversarial_meta_sampling}. Other approaches use
phoneme overlap and pronunciation information for dynamic multilingual
sampling~\cite{dynamic_sampling_meta_asr}, or estimate positive
cross-lingual transfer through acoustic pseudo-tokens and clip-level
acoustic similarity~\cite{acoustic_pseudotokens,catds}. These findings
show that data quantity or loss alone cannot fully determine the
usefulness of multilingual speech.

Beyond multilingual ASR, recent studies have investigated the use of LLMs and agentic mechanisms for data selection and preparation.
AskLLM leverages the zero-shot reasoning capability of an instruction-tuned LLM to directly assess the quality of individual training examples, demonstrating that LLM-based semantic judgments can serve as effective data-selection signals~\cite{sachdeva2024train}. Bai et al.\ introduce a multi-actor collaborative framework in which multiple data-selection strategies independently prioritize samples and a central console dynamically combines their decisions according to the current training state~\cite{bai2025efficient}. Moving beyond sample scoring, DataFlow introduces an LLM-driven agent that translates natural-language data requirements into executable preparation pipelines composed of generation, evaluation, filtering, and refinement operators~\cite{liang2025dataflow}.

These agentic data sampling method has not been adapted to multilingual ASR. Existing approaches, however, generally employ a predefined selection
criterion and operate at one granularity, such as language weighting or
utterance selection. Coordinating language-level allocation,
utterance-level relevance, and token-level supervision according to
changing multilingual ASR errors remains comparatively underexplored\cite{peng2025survey}.

\section{Method}
\begin{figure}[t]
\centering
\makebox[\textwidth][c]{%
\includegraphics[width=\textwidth]{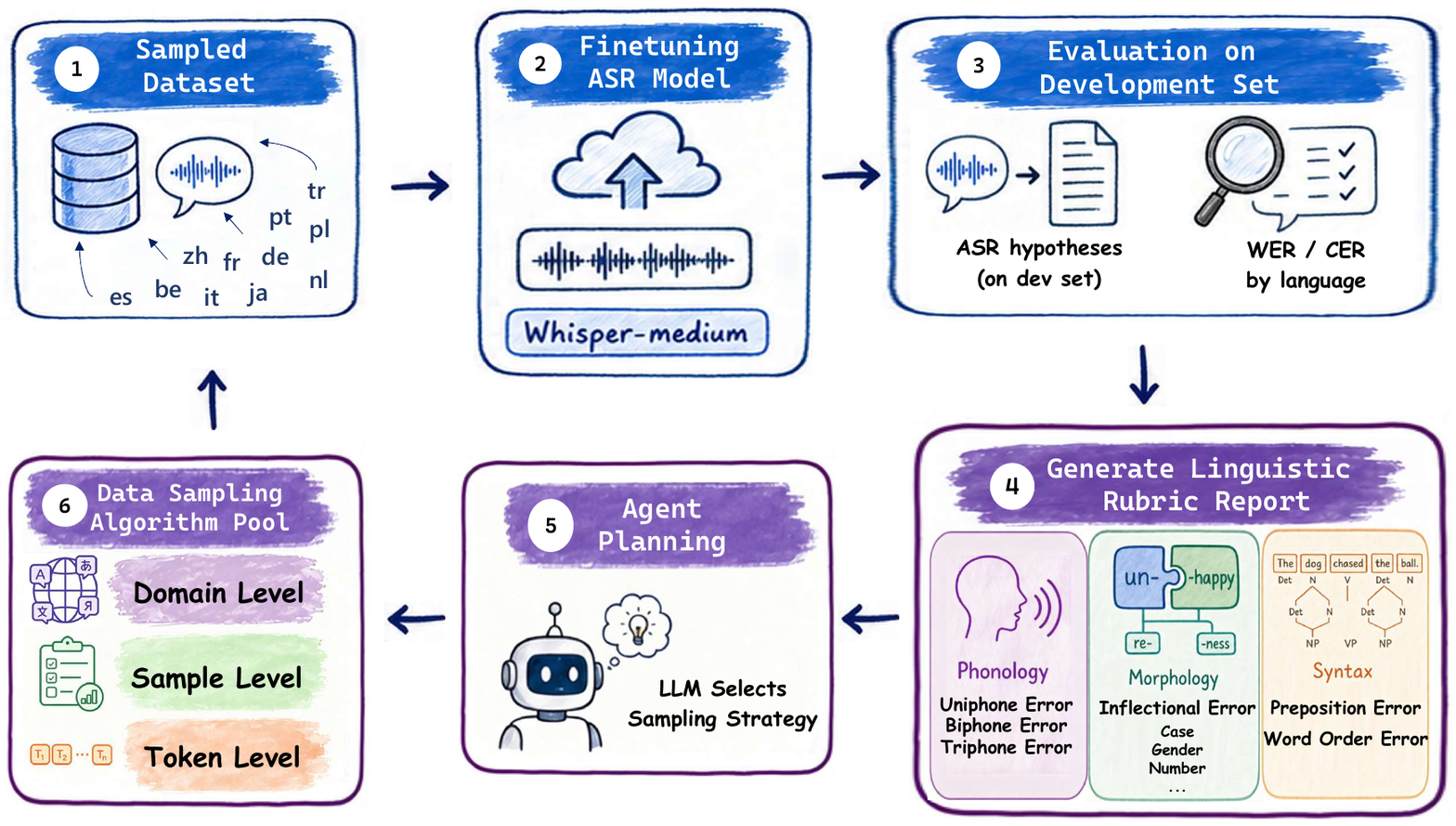}%
}
\caption{Overview of the proposed agentic data sampling framework. 
The blue nodes show the common ASR model training pipeline, while the purple nodes belong to our agentic data sampling workflow.}
\label{fig:framework}
\end{figure}

We propose a linguistic-aware agentic data sampling framework. The system use an LLM-based agent to drive the data sampling process to train a multi-lingual ASR model.  
The overall pipeline is shown in Fig.~\ref{fig:framework}.

In summary, the agentic workflow proceeds as an iterative training loop. At each epoch, the system first samples a batch from the training corpus and uses it to fine-tune the ASR model. The fine-tuned model is then evaluated on the development set, and its recognition outputs are converted into a linguistic rubric report. Based on this report together with the training statistics, the LLM agent designs a three-stage data-sampling plan covering domain-level, sample-level, and token-level decisions. This plan is then applied to the training corpus to construct the subset used for fine-tuning in the next epoch.

\subsection{Linguistic Rubric}
We propose a linguistic rubric for our multilingual ASR data sampling agent, which converts raw ASR recognition errors into more detailed statistics from a linguistic perspective, and provides structured feedback for data sampling. 
The rubric attributes some ASR errors into three linguistic dimensions: phonology, morphology, and syntax. 
The goal of the rubric is not to replace WER or CER, but to explain which linguistic phenomena contribute to the observed errors.
Based on the detailed error distribution in each linguistic dimension, the agent can identify more specific weaknesses and select more targeted sampling strategies.

\paragraph{Phonology.}
The phonology dimension analyzes ASR errors at the phoneme level. We convert grapheme-level references and predictions into IPA representations using eSpeak NG~\cite{espeakng}, and then identify errors at the granularity of uniphones, biphones, and triphones.
This dimension helps identify which sounds or sound combinations are difficult for the model and supports phone-aware sample selection.

\paragraph{Morphology.}
The morphology dimension focuses on word-formation errors, especially inflectional errors and case, gender, number, or tense marker errors.
We compare aligned reference and hypothesis tokens to determine whether the main lexical stem is preserved while a morphological marker changes or disappears. In morphologically rich languages, a single inflectional error can alter the grammatical or semantic function of a word while causing only a minimal change in character-level edit distance.

\paragraph{Syntax.}
The syntax dimension focuses on errors involving prepositions, function words, and local word-order patterns. Although ASR is an acoustic-to-text task, recognition errors often appear as systematic textual mistakes, such as dropped function words or confused short grammatical words. Syntax-level analysis therefore provides additional guidance for selecting utterances that contain error-prone grammatical contexts.

For each linguistics dimension, the rubric report contains both categorical distribution and short natural-language summaries. The report is then passed to the agent together with WER/CER and training statistics. In this way, the agent can distinguish, for example, a language whose high WER is mainly caused by phoneme confusions from another language whose errors are dominated by inflectional endings or function words.

\subsection{Data Sampling Agent}

\subsubsection{Hierarchical Algorithm Pool}
We formulate data sampling for multilingual ASR as a hierarchical decision process with three stages: domain, sample, and token. These stages correspond to three increasingly fine-grained data sampling level: the language ratios of the training dataset, the utterances to be selected for each language, and operations (e.g. gradient weights or masking) applied to tokens during optimization. 

The agent designs the data sampling plan using a hirachical algorithm pool. 
At each training epoch, the agent observes the linguistic rubric report, training statistics, and budget constraints, then selects one compatible algorithm from each level together with its parameters. The algorithms selected by the data sampling agent are executed in hierarchical order. 

\begin{table}
\caption{Multi-level data sampling algorithm pool.}
\label{tab:algorithm_pool}
\centering
\small
\begin{tabularx}{\textwidth}{@{}llX@{}}
\toprule
Algorithm & Stage & Description \\
\midrule
Doremi & Domain & Updates language sampling weights using loss-based signals. \\
Exp-Reweight & Domain & Reweights languages using an agent-selected score. \\
DOGE & Domain & Uses gradient similarity for domain-level weighting. \\
\midrule
LESS & Sample & Selects utterances using training-validation gradient alignment. \\
Phone Error Distribution & Sample & Selects utterances matching phoneme error distributions. \\
\midrule
RHO-1 & Token & Selects informative tokens using token-level loss or entropy. \\
TOSS & Token & Filters tokens according to utility or risk scores. \\
\bottomrule
\end{tabularx}
\end{table}

We predefined the algorithm pool using 7 algorithms as shown in Table~\ref{tab:algorithm_pool}. 
The pool is extensible: each algorithm is a self-contained module that only requires its declared inputs to execute. Once implemented, registering a new algorithm simply requires providing a profile describing its behaviour and parameter schema. The agent can then reason about it based on its profile alongside all other registered algorithms.

\subsubsection{Data Sampling Procedure}
The decision-making process of agentic data sampling can be formulated as follows. Given the original multilingual corpus $\mathcal{D}$, the agent selects a sequence of three actions:

\begin{equation}
a = (a^{domain}, a^{sample}, a^{token}),
\end{equation}

where each action is itself a pair $a^{l} = (m^{l}, \theta^{l})$ consisting of an algorithm choice $m^{l}$ from a level of the predefined pool and its parameters $\theta^{l}$. The actions are applied sequentially:

\begin{equation}
\mathcal{S} = a^{token}(a^{sample}(a^{domain}(\mathcal{D}))).
\end{equation}

which first adjusts language-level sampling weights, then selects utterances under the adjusted weights, and finally applies per-token operation during optimization.

At each data-sampling stage, the agent receives a compact state description that combines the current epoch statistics, including the training-loss curve and WER/CER with per-language breakdowns, with the linguistic rubric reports. This state also includes the available algorithm profiles and their parameter schemas, the execution results from earlier sampling stages in the same epoch, such as the selected algorithm, parameters, and filtered dataset, and the running history of the agent's decisions in previous epochs. Given this information, the agent outputs a structured decision for the current stage, specifying the selected algorithm and parameters together with an interpretable rationale for the choice.









\section{Experiments}
\subsection{Setup}
\subsubsection{Dataset}
We employ a 12-language subset of Common Voice~\cite{commonvoice} as our benchmark, covering  Belarusian (be), Chinese (zh), Dutch (nl), French (fr), German (de), Italian (it), Japanese (ja), Polish (pl), Portuguese (pt), Russian (ru), Spanish (es), and Turkish (tr). This subset comprises a training corpus, a development set, and a test set. Both the development and test sets are 12 hours, equally distributed with exactly one hour per language.
The training corpus comprises a total duration of 600 hours, with each language contributing an equal share of 50 hours. 

\subsubsection{Baselines and Implementation}

We fine-tune the Whisper-medium~\cite{whisper} model on this dataset using various data sampling strategies to evaluate their impact on automatic speech recognition (ASR) performance. To isolate the effects of the sampling strategies, all methods are assessed under identical model architectures, optimizer configurations, and evaluation protocols. Specifically, we employ Low-Rank Adaptation~\cite{lora} with a rank $r=64$, $\alpha=64$, and a dropout rate of 0.05. The model is optimized using the AdamW optimizer with a learning rate of $1\times10^{-4}$ and a per-device batch size of 128, distributed across 8 Ascend 910B NPUs.

To ensure a fair comparison, the total duration of trained data for all models is strictly fixed at 1,200 hours. We benchmark our proposed agentic data sampling method against two baselines: a no-sampling baseline and a random sampling baseline. The no-sampling baseline utilizes the complete training corpus without any selection strategy; specifically, it was trained on the full 600-hour data for 2 epochs. The random sampling method involves randomly selecting 300 hours of data from the training corpus, repeated over 4 iterations. For our agentic sampling method, we evaluated different sampling durations, extracting batches of 100 and 300 hours of data from the training corpus respectively to reach the target training volume. Consequently, the agentic system performs 12 and 4 sampling iterations respectively.

\subsection{Main Results}
\begin{figure}[h]
\centering
\includegraphics[width=\textwidth]{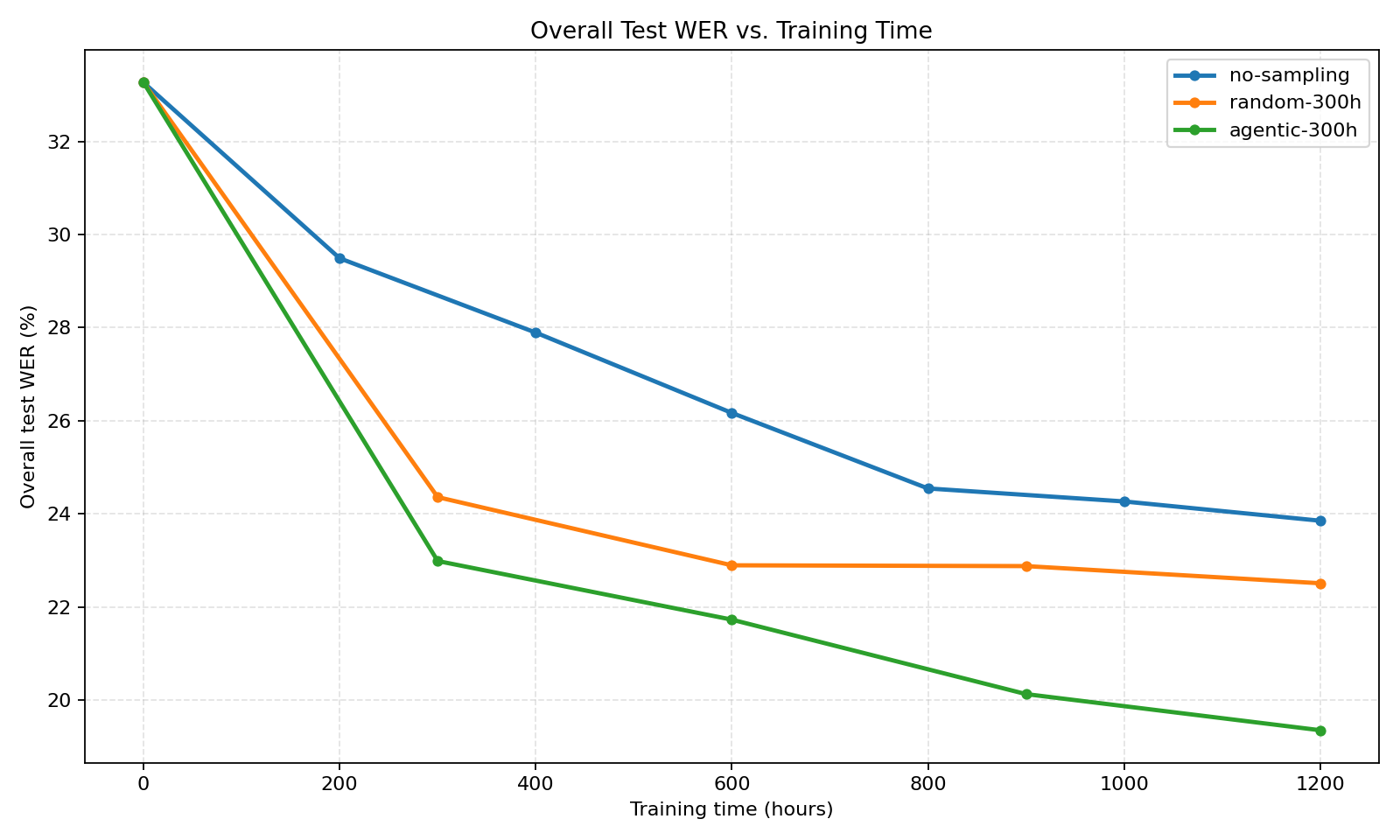}
\caption{Comparison of overall test WER over training time under no sampling, random sampling, and agentic sampling. The sampling batch of random and agentic method is 300 hours.}
\label{fig:test_wer}
\end{figure}
\begin{table}
\caption{Comparison of ASR performance across different data sampling strategies. The suffix of each method denotes the data duration sampled per iteration. Performance is measured in Character Error Rate (CER) for Japanese (ja) and Chinese (zh), and Word Error Rate (WER) for all other languages. The best results for each language are highlighted in bold.}
\label{tab:main_result}
\centering
\makebox[\textwidth][c]{%
\resizebox{\textwidth}{!}{%
\begin{tabular}{lccccccccccccc}
\toprule
Method & be & de & es & fr & it & ja & nl & pl & pt & ru & tr & zh & Avg. \\
\midrule
no-sampling & 70.3 & 39.9 & 20.5 & 21.5 & 10.9 & 10.4 & 19.0 & 21.0 & 18.7 & 19.2 & 17.3 & 22.3 & 23.9 \\
random-300h & 67.6 & 36.1 & 18.6 & 20.1 & 10.1 & \textbf{9.1} & 18.3 & 19.0 & 17.9 & 15.6 & 20.2 & \textbf{20.4} & 22.5 \\\hline
agentic-100h & 75.9 & \textbf{28.7} & 19.2 & 17.4 & 8.9 & 10.1 & 14.4 & \textbf{15.5} & 12.4 & \textbf{13.9} & 15.6 & 24.1 & 21.1 \\
agentic-300h & \textbf{66.2} & 30.1 & \textbf{12.6} & \textbf{17.1} & \textbf{8.6} & 9.7 & \textbf{14.0} & 15.7 & \textbf{11.6} & 15.4 & \textbf{14.4} & 20.5 & \textbf{19.4} \\
\bottomrule
\end{tabular}
}
}
\end{table}

Table~\ref{tab:main_result} presents the comparative analysis of ASR performance across models trained with varying data sampling strategies. 
Our proposed agentic sampling strategies demonstrate superior performance compared to the baselines. Specifically, the agentic sampling method with 300 hours sampling batch achieves the best average performance with an error rate of 19.4, significantly outperforming the no-sampling baseline (23.9) and the random sampling baseline (22.5).
In terms of specific language performance, it yields the lowest error rates in 9 out of 12 languages and shows significant improvement in Indo-European languages, especially Romance languages. The agentic-300h method reduced the WER for Spanish and Portuguese by 38.5\% and 38.0\% respectively, compared with the no-sampling baseline. This may be benefit from our linguistic-aware rubric as these languages shares similar linguistic features.

As shown in Figure~\ref{fig:test_wer}, the baseline systems tend to converge after approximately 1200 hours of training, whereas the agentic sampling method continues to reduce the overall WER. This suggests that agentic sampling selects more informative training data and remains effective beyond the point where the baseline strategies begin to saturate.

\subsection{Case Analysis}

\begin{table*}[t]
\centering
\small
\caption{Representative test cases corrected by agentic data sampling. Red text marks persistent recognition errors, while green text marks the corrected linguistic unit after finetuning.}
\label{tab:error_case_study}
\setlength{\fboxsep}{5pt}
\begin{tabular}{@{}p{\textwidth}@{}}
\fbox{\parbox{0.97\textwidth}{%
\textbf{Morphology} \hfill French\\[-1mm]
\rule{\linewidth}{0.3pt}
\textbf{Reference:} yvan publie des anthologies de poètes allemands et français et vit de ses traductions\\[1mm]
\textbf{Whisper:} \textcolor{red}{yvon} publie des anthologies de poètes \textcolor{red}{allemand} et français et vit de \textcolor{red}{ces} traductions\\[1mm]
\textbf{No-sampling:} \textcolor{red}{yvon} publie des anthologies de poètes \textcolor{red}{allemand} et français et vit de ses traductions\\[1mm]
\textbf{Agentic:} \textcolor{green!50!black}{yvan} publie des anthologies de poètes \textcolor{green!50!black}{allemands} et français et vit de ses traductions
}}\\[2mm]
\fbox{\parbox{0.97\textwidth}{%
\textbf{Syntax} \hfill Italian\\[-1mm]
\rule{\linewidth}{0.3pt}
\textbf{Reference:} a causa degli enormi debiti fu costretta a vivere grazie a dei prestiti\\[1mm]
\textbf{Whisper:} a \textcolor{red}{kaoza da gdje normi} debiti \textcolor{red}{fukustreta} a vivere \textcolor{red}{gracija da ih} prestiti\\[1mm]
\textbf{No-sampling:} a \textcolor{green!50!black}{causa degli enormi} debiti \textcolor{green!50!black}{fu costretta} a vivere grazie \textcolor{red}{ai} prestiti\\[1mm]
\textbf{Agentic:} a \textcolor{green!50!black}{causa degli enormi} debiti \textcolor{green!50!black}{fu costretta} a vivere \textcolor{green!50!black}{grazie a dei} prestiti
}}\\[2mm]
\fbox{\parbox{0.97\textwidth}{%
\textbf{Phonology} \hfill Portuguese\\[-1mm]
\rule{\linewidth}{0.3pt}
\textbf{Reference:} ivaí\\[1mm]
\textbf{Whisper:} \textcolor{red}{e vai}\\[1mm]
\textbf{No-sampling:} \textcolor{red}{e vai}\\[1mm]
\textbf{Agentic:} \textcolor{green!50!black}{ivaí}
}}
\end{tabular}
\end{table*}

Table~\ref{tab:error_case_study} presents three representative cases of errors in each linguistic dimension. The base Whisper model makes the error and no-sampling finetuning still fails to correct it, while our agentic sampling model produces the correct transcription. 
These examples illustrate that the improvement of agentic data sampling is not limited to general lexical recovery, but also appears in linguistically meaningful error categories.

In the morphology case, an inflectional error occurs in the baseline as it transcribes \textit{poètes allemands} as \textit{poètes allemand}, dropping the plural agreement marker \textit{-s} on the adjective. 
The agentic model restores \textit{allemands}, suggesting that the selected training samples help the model better preserve inflectional information in Romance languages. 

In the syntax error case, the baseline simplifies the prepositional phrase \textit{grazie a dei prestiti} to \textit{grazie ai prestiti}. The agentic model, however, recovers the full prepositional construction, demonstrating an improved ability to model function-word structures rather than merely correcting surface-level spelling errors. This enhancement is likely attributable to token-level data sampling, which effectively resolves uncertainty at the local token level.


The phonology case shows a clear diagnosis-action-test loop together with the rubric traces. This phone error is identified in the rubric, and the phone error distributiont sampling algorithm is selected in the middle epochs to target such phonological patterns, and the final model resolves a persistent phonological confusion on the test set. 

Overall, these cases suggest that agentic data sampling improves ASR behavior across morphology, syntax, and phonology by selecting training data that is more closely aligned with the model's observed error patterns.

\section{Analysis}
\label{sec:analysis}

\subsection{Agent Decision Behavior}
\label{sec:agent-decision-behavior}

We further examine whether the sampling policy produced by the agent is
adaptive and interpretable rather than equivalent to a fixed schedule.
At each round, the agent receives the evaluation history and linguistic
error diagnostics from the preceding round, and jointly selects
domain-, sample-, and token-level sampling strategies for the next
round.

\begin{table}[t]
\caption{Representative agent-selected sampling strategies.}
\label{tab:agent_behavior}
\centering
\small
\begin{tabularx}{\textwidth}{@{}c>{\raggedright\arraybackslash}Xll@{}}
\toprule
Round & Main Error Pattern & Domain Action & Sample/Token Action \\
\midrule
1 & Cold-start uncertainty without reliable error feedback
  & DoReMi & LESS + RHO-1 \\
2 & Concentrated phonological errors
  & Exp-Reweight & PE-DSTB + TOSS \\
3 & Persistent phone-level confusions
  & DOGE & PE-DSTB + RHO-1 \\
4 & Language-specific phonetic errors
  & DoReMi & LESS + TOSS \\
\bottomrule
\end{tabularx}
\end{table}

Table~\ref{tab:agent_behavior} shows a transition from a conservative
cold-start policy to diagnosis-conditioned decisions. In the first
round, no previous evaluation history is available, so the agent selects
broadly applicable loss-based methods rather than making
language- or error-specific interventions. Once diagnostic feedback
becomes available, the selected actions change across rounds instead of
repeating the initial configuration. This behavior indicates that the
policy is conditioned on the evolving error profile of the ASR model
rather than following a predefined sampling schedule.

More importantly, the selected actions are aligned with the granularity
of the observed errors. When phonological errors become concentrated,
the agent introduces \textsc{PE-DSTB} at the sample level, connecting
phone-level diagnostics to the selection of training utterances.
Meanwhile, the alternation among \textsc{Exp-Reweight}, \textsc{DOGE},
and \textsc{DoReMi} suggests that the agent distinguishes between
multilingual allocation and within-domain data quality. Domain-level
actions determine where training emphasis should be placed, whereas
sample- and token-level actions determine which supervision within
those domains is most useful.

The alternation between \textsc{RHO-1} and \textsc{TOSS} further shows
that the agent does not treat all difficult tokens as equally valuable.
Some persistent errors may require greater emphasis on under-learned
tokens, while others may benefit from filtering unreliable token-level
signals. The resulting policy is therefore hierarchical: the agent
first identifies the data regions requiring intervention and then
selects an appropriate level of filtering or emphasis within those
regions.

Overall, the decisions provide evidence that the agent uses linguistic
diagnostics to revise both the target and the granularity of data
selection. These examples do not establish that every selected strategy
is globally optimal, but they demonstrate that the decision process is
traceable, responsive to model feedback, and distinct from a fixed
combination of sampling algorithms. Together with the linguistic error
reductions in Table~\ref{tab:error_reduction}, this behavior supports the
interpretation of the proposed framework as an adaptive feedback loop
between error diagnosis and data selection.



\subsection{Linguistic Error Reduction}
\label{sec:linguistic-error-reduction}

Beyond aggregate WER/CER, we further examine whether agentic data
sampling reduces linguistically meaningful recognition errors. We
compare the original Whisper-medium model with the model obtained
after four epochs of agentic data sampling. Both models are evaluated
on the same Common Voice test set covering 12 languages. For
phonological analysis, we convert each reference and hypothesis into
IPA phone sequences using G2P, strip non-phonemic markers, and count
all mismatched 1-, 2-, and 3-phone $n$-gram events over the full test
set. For morphology and syntax, we use rule-based detectors for
inflectional marker errors, preposition/function-word errors, and
word-order deviations. The relative reduction is computed with
respect to the original Whisper-medium model.

\begin{table}[t]
\caption{Linguistic error counts of the original
and model trained with agentic data sampling.}
\label{tab:error_reduction}
\centering
\small
\begin{tabular}{llrrr}
\toprule
Category & Error Type & Original & Agentic-trained & Reduction \\
\midrule
Phonology
& Uniphone 
& 124,191 & 114,831 & 9,360 (7.5\%) \\

& Biphone 
& 129,700 & 121,123 & 8,577 (6.6\%) \\

& Triphone 
& 131,243 & 122,970 & 8,273 (6.3\%) \\

\midrule
Morphology
& Inflectional marker / word-ending
& 975 & 467 & 508 (52.1\%) \\

\midrule
Syntax
& Preposition / function-word
& 526 & 403 & 123 (23.4\%) \\

& Word-order deviation
& 3 & 1 & 2 (66.7\%) \\
\bottomrule
\end{tabular}
\end{table}

Table~\ref{tab:error_reduction} reveals a complementary pattern across
linguistic levels. Phonological errors contribute the largest absolute
reductions, reflecting their high frequency in multilingual ASR, while
morphological errors exhibit the largest relative improvement. This
suggests that the benefit of agentic data sampling is not confined to
frequent acoustic--phonetic confusions: it also extends to less frequent
but linguistically informative errors, such as inflectional suffixes and
word-final grammatical markers. The consistent reductions from
uniphone to triphone errors further indicate improved modeling of both
individual phonemes and short phonetic contexts, although the smaller
gain at higher $n$-gram orders suggests that longer context-dependent
confusions remain more difficult.

These results support the central motivation of our agentic sampling
framework. A uniform sampling strategy treats all training data as
equally useful, whereas the agent adjusts the data distribution using
historical model performance and linguistic error feedback. The observed
error profile is consistent with the agent allocating additional
training exposure to languages or samples containing persistent
phonological contrasts, morphologically marked word forms, and
grammatical function words. In particular, the substantial reduction in
morphological errors suggests that adaptive sampling can improve
phenomena that may be underrepresented in the overall training
distribution and therefore receive insufficient attention under uniform
sampling.

The reduction in preposition and function-word errors also indicates
that the improvement is not limited to content-word recognition, but
extends to local grammatical information. However, the number of
word-order deviations is too small to draw a reliable conclusion about
global syntactic modeling. Overall, the linguistic breakdown provides
evidence that agentic data sampling changes not only the aggregate error
rate but also the composition of the remaining errors, yielding
improvements across acoustic--phonetic and grammatical dimensions that
cannot be fully characterized by WER/CER alone.





\section{Conclusion}
\enlargethispage{2\baselineskip}

We proposed a linguistic-aware agentic data sampling framework for
multilingual ASR that converts phonological, morphological, and syntactic
error diagnostics into domain-, utterance-, and token-level sampling
decisions. By placing evaluation inside the training loop, the framework
treats data sampling as an adaptive process rather than a fixed
preprocessing strategy.

Experiments on twelve Common Voice languages with Whisper-medium show
that the agentic-300h setting achieves an average error rate of 19.4,
compared with 23.9 for no sampling and 22.5 for random sampling, and
obtains the best result on nine languages. The method also reduces
phone-level, inflectional, and function-word errors, while the selected
strategies change with the evolving model error profile. These results
suggest that multilingual ASR evaluation can serve not only to measure
performance, but also to provide actionable feedback for subsequent data
selection. Future work will extend the framework to broader language
coverage and conduct finer-grained ablations of the linguistic rubric
and individual sampling decisions.


%
%
\newpage
\bibliographystyle{splncs04_unsrt}
\bibliography{reference}

\newpage
\section*{Appendix 1}
Structure of agent decision in each stage:
\begin{verbatim}
{
  "rationale": {
    "algorithm": "Per-language CER shows high variation...",
    "param": "alpha=0.8 to smooth weights for low-resource languages..."
  }
  "algorithm": "doremi",
  "params": {"alpha": 0.8, "reweight_eta": 2.0, ...},
}
\end{verbatim}
\section*{Appendix 2}

Example of linguistic rubric:
\begin{verbatim}
{
  "phonology": {
    "uniphone_errors": [
      {"phone": "i", "error_rate": 0.68, "languages": ["be", "pl", "pt"]},
      ...
    ],
    "biphone_errors": [...],
    "triphone_errors": [...],
    "summary": "High phonological error rates are observed..."
  },
  "morphology": {
    "inflection_errors": [
      {"language": "de", "marker": "er", "error_rate": 0.10, ...}
      ...
    ],
    "summary": "Inflectional errors concentrate on..."
  },
  "syntax": {
    "prep_errors": [
      {"language": "fr", "prep": "de", "error_rate": 0.13, ...}
      ...
    ],
    "summary": "Preposition and function-word errors are frequent..."
  }
}
\end{verbatim}

\section*{Appendix 3: Prompt Template}
The prompt template used by the LLM agent at each sampling stage is as follows. Placeholders enclosed in braces are filled by the workflow before each LLM call.
\begin{PromptBlock}
You are a data sampling agent for multilingual ASR training.

## Context

The pipeline runs three stages sequentially:
  domain (= language) -> sample (= utterance) -> token (= subword)

Each stage's output feeds into the next.

## Current stage

You are configuring **Stage {stage_num} -- {stage_name} ({stage_level})**.
Available algorithms: {available_algorithms}
This stage runs **{stage_phase}**.

## What you will receive

- ``rubric_diagnosis``: error patterns, per-language CER, overall CER,
  phone/grapheme distributions, and summary text.
- ``previous_stages``: summary of what earlier stages did and their results.
  (empty for the first stage)
- ``selection_context``: experiment-level constraints and cross-epoch
  algorithm usage history, including under-used algorithms and recent
  domain->sample->token paths.

## Rules

1. Choose exactly ONE algorithm from the available list.
2. Set specific parameter values using the parameter schemas below.
3. Prefer conservative settings unless the diagnosis clearly supports aggressive filtering.
4. This experiment must explore the algorithm pool across epochs. Prefer
   under-used algorithms and new stage combinations when their required
   inputs are available from the rubric or from orchestrator precompute.
   If an algorithm has zero or much lower usage than the alternatives at the
   same stage, choose it before reusing a heavily repeated stage algorithm
   unless it is invalid for the current inputs.
   Avoid repeating the same stage algorithm for more than two consecutive
   epochs unless every alternative is invalid; if you repeat, explain why.
5. Do not choose an algorithm only because it was used in the previous epoch.
   Balance current diagnosis fit with coverage of doremi/exp_reweight/doge,
   less/pe_dstb, and rho1/toss over the full run.
6. Return valid JSON matching the required schema.
7. Output schema fields following their original order in the schema (rationale -> algorithm -> params).

.. note::
   File-path parameters such as ``error_results_path``, ``loss_path``, and
   ``score_path`` must refer to real run artifacts when you set them.  For
   model-based score fields such as DoReMi losses or rho1 token scores, the
   orchestrator precomputes missing fields before training from the current
   epoch checkpoint and the fixed train split.  Algorithms do not fall back
   to package-local offline scored JSONL files.

## Algorithm parameter schemas

{algo_schemas}
\end{PromptBlock}

\end{document}